\documentclass[runningheads]{llncs}
\usepackage{accvabbrv}
\usepackage[T1]{fontenc}
\usepackage{appendix}
\RequirePackage{subcaption}
\usepackage{graphicx}
\usepackage{booktabs}
\usepackage{algorithm,algpseudocode}
\usepackage{bbding}
\usepackage{array}
\usepackage[accsupp]{axessibility}
\usepackage[pagebackref,breaklinks,colorlinks]{hyperref}
\begin{document}
\title{TaE: Task-aware Expandable Representation for Long Tail Class Incremental Learning}
\titlerunning{Task-aware Expandable Representation for LT-CIL}
%
\author{Linjie Li\inst{1} \and
Zhenyu Wu\inst{1}\thanks{Corresponding author} \and
Jiaming Liu\inst{2}\and
Yang Ji\inst{1}
}
\authorrunning{Li L.  et al.}
%
\institute{School of Information and Communication Engineering, Beijing University of Posts and Telecommunications, Beijing, 100876, China \and
School of Computer Science, Peking University, Beijing, 100871, China
\\
\email{shower0512@bupt.edu.cn}}
\maketitle              
\begin{abstract}
Class-incremental learning is dedicated to the development of deep learning models that are capable of acquiring new knowledge while retaining previously learned information. Most methods focus on balanced data distribution for each task, overlooking real-world long-tailed distributions. Therefore, Long-Tailed Class-Incremental Learning has been introduced, which trains on data where head classes have more samples than tail classes. Existing methods mainly focus on preserving representative samples from previous classes to combat catastrophic forgetting. Recently, dynamic network algorithms freeze old network structures and expand new ones, achieving significant performance. However,  with the introduction of the long-tail problem, merely extending Determined blocks can lead to miscalibrated predictions, while expanding the entire backbone results in an explosion of memory size. To address these issues, we introduce a novel \textbf{T}ask-\textbf{a}ware \textbf{E}xpandable (\textbf{TaE}) framework, dynamically allocating and updating task-specific trainable parameters to learn diverse representations from each incremental task while resisting forgetting through the majority of frozen model parameters. To further encourage the class-specific feature representation, we develop a \textbf{C}entroid-\textbf{E}nhance\textbf{d} (\textbf{CEd}) method to guide the update of these task-aware parameters. This approach is designed to adaptively allocate feature space for every class by adjusting the distance between intra- and inter-class features, which can extend to all "training from sketch" algorithms. Extensive experiments demonstrate that TaE achieves state-of-the-art performance.

\end{abstract}

\section{Introduction}

\begin{figure}[t]
\centering
\includegraphics[width=1\textwidth]{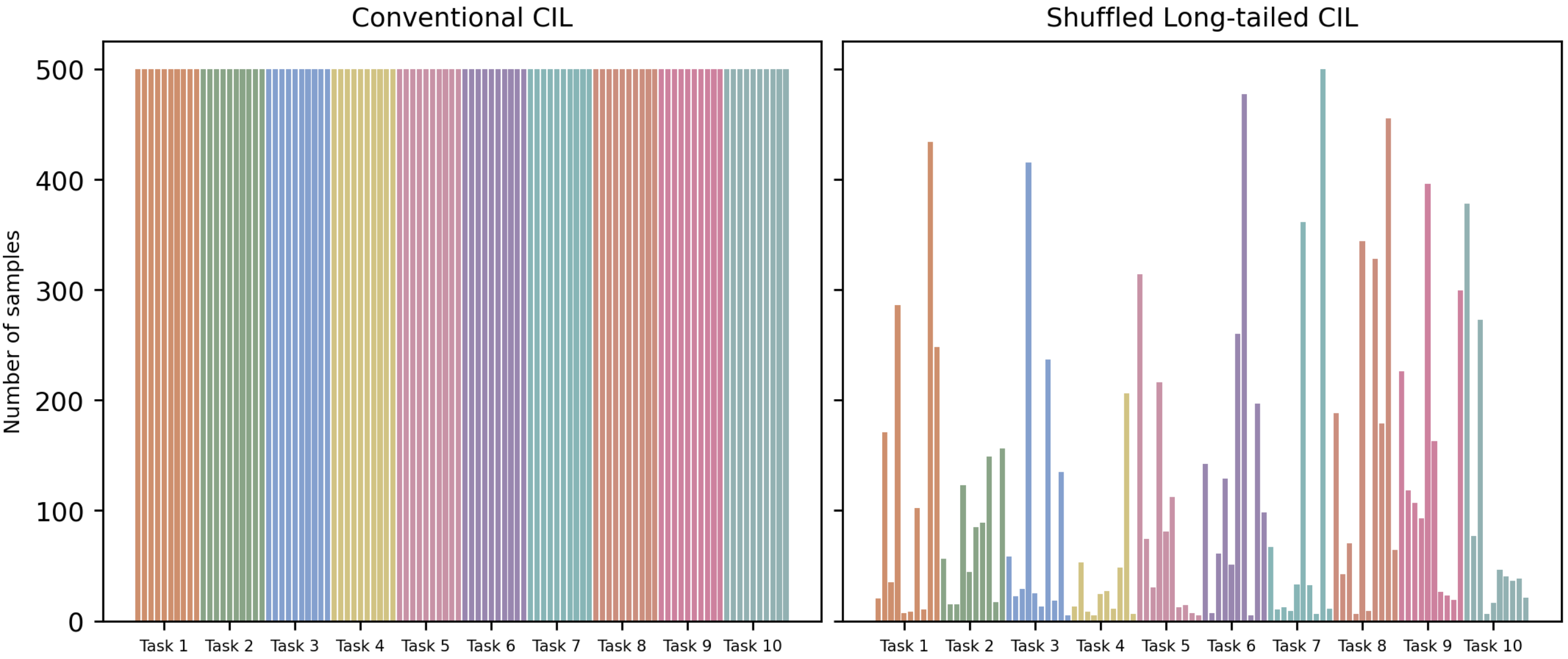}
\caption{Illustration of Conventional and Shuffled distribution.}
\label{distribution}
\end{figure}

Class Incremental Learning (CIL) aims to establish a unified classifier across all known classes, and the most challenging problem of CIL is catastrophic forgetting \cite{catastrophic,dilmamaCVPR,CILsurveyTPAMI}. Conventional CIL makes the assumption that a balanced distribution forms the training set. On the other hand, real-world data distributions are typically long-tailed \cite{DeepLTsurvey,rehearsal}, meaning that there are significantly more head-class samples than tail-class samples. Proposals have recently been made for Long-tailed Class-incremental learning (LT-CIL)\cite{LTCIL}. Shuffled LT-CIL biases the model toward learning head classes while neglecting tail classes during new task training, as illustrated in \ref{distribution}. This leads to a deficiency in the learning of the tail class features. Similarly, when retaining old knowledge, the model experiences more severe forgetting due to the blurring of tail class features. As such, the performance of the CIL approach is decreased by this long-tailed specialized catastrophic forgetting\cite{LTCIL}.

Currently, dynamic networks exhibit commendable performance in both CIL and LT-CIL \cite{deepCILsurvey,DER,memo,beef,expert,LTCIL}. In LT-CIL tasks, the dynamic network's ability to expand its structure enhances the model's representative capacity, allowing it to better adapt to the data distribution of new tasks. As a result, the model demonstrates stronger learning capabilities in handling long-tailed distributions within new tasks. Dynamic network approaches can be roughly divided into two categories: expanding a part of the network structure, like deep blocks\cite{memo}, or fully expanding the backbone\cite{DER,Foster}. However, due to the long-tailed distribution, shallow networks' ability to represent is not as general as a balanced distribution. Therefore, only expanding the deep blocks in every task and freezing shallow blocks might lead to miscalibrated predictions. Conversely, fully expanding the entire backbone could result in an explosion of memory size.

\begin{figure}[t]
\centering
\includegraphics[width=0.6\textwidth]{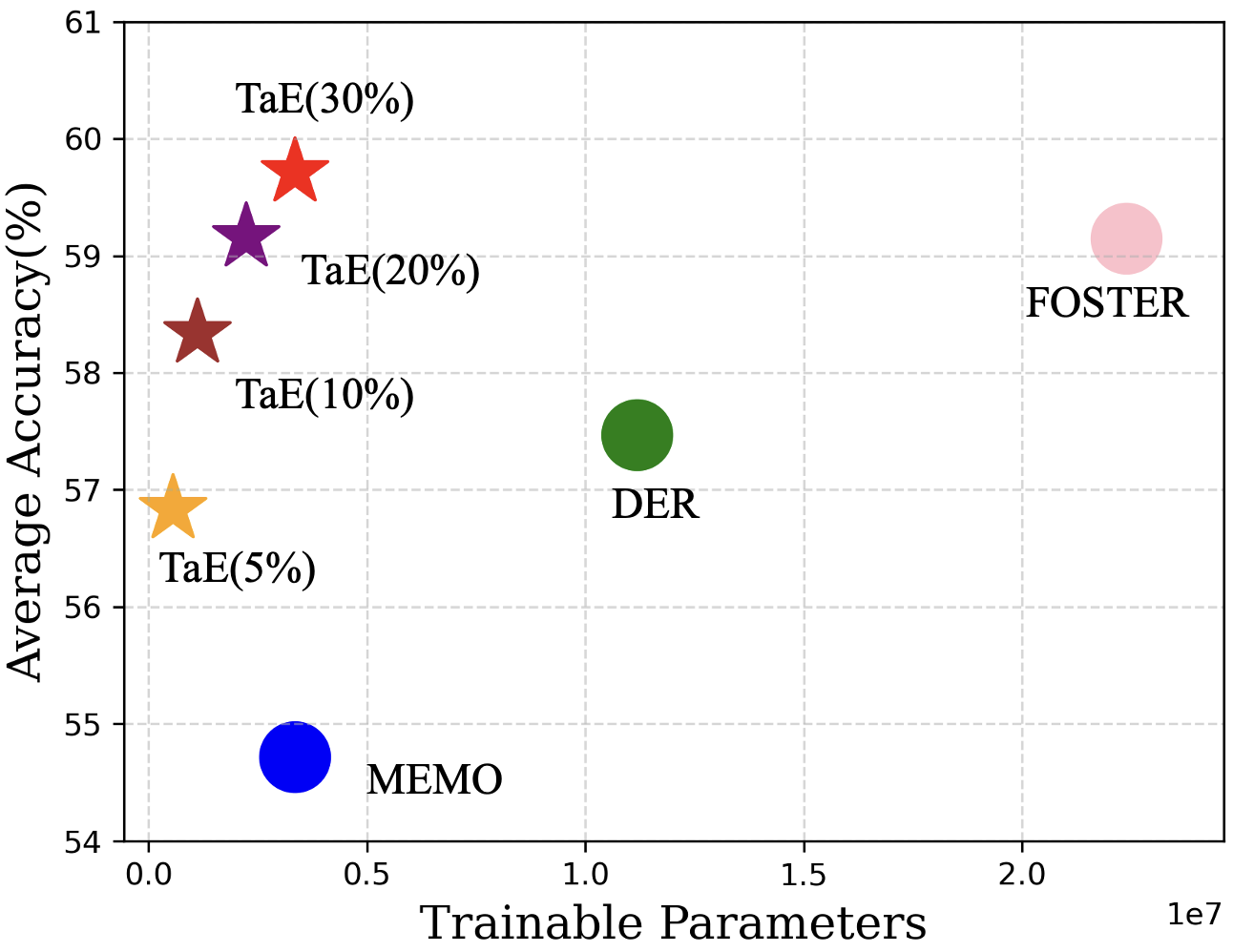}
\caption{Parameter-performance comparison of different dynamic network methods on ImageNet100-LT B0-10steps. TaE only expands a few training parameters to exceed the SOTA CIL method.}
\label{motivation}
\end{figure}

In this paper, we introduce a novel \textbf{T}ask-\textbf{a}ware \textbf{E}xpandable (TaE) framework, dynamically allocating and updating task-specific trainable parameters to learn diverse representations from each incremental task, while resisting forgetting through the majority of frozen model parameters. Specifically, before training on a new task, we conduct forward propagation for each sample and accumulate the entire parameter gradients to identify the top-$p$$\%$(eg,.5$\%$, 10$\%$, 20$\%$, 30$\%$) most sensitive parameters. These parameters are then exclusively updated during training. To further enhance class-specific feature representation, we develop a \textbf{C}entroid-\textbf{E}nhance\textbf{d} (CEd) method to guide the update of task-aware parameters. We acquire a centroid for each class and maintain a set of centroids. These centroids are treated as learnable parameters, dynamically updated throughout the model training process to adapt to changes in the feature space resulting from the introduction of new data. This approach aims to adaptively minimize intra-class feature distances while maximizing inter-class feature distances among all observed classes. The CEd method enhances the class-specific representation of task-aware parameters, which decreases the overfitting of head classes and underfitting of tail classes.

We conducted experiments on two commonly used benchmark datasets, including CIFAR100 and ImageNet100. We set both datasets with a long-tail configuration and shuffled their order. We set up different ratios of long-tail and different ways of incrementation. Extensive experiments and ablation studies prove the effectiveness of our method. Especially in representative experiments on the Imagenet100 dataset, an extension by just 5$\%$ results in surpassing MEMO's performance by 0.64$\%$ in final accuracy and 2.12$\%$ in average accuracy. Furthermore, an extension by 10$\%$ leads to surpassing DER with improvements of 1.20$\%$ and 0.87$\%$ in final accuracy and average accuracy, respectively, as depicted in \ref{motivation}.
The main contributions can be summarized in three points:
\begin{itemize}
    \item  We introduce a novel Task-aware Expandable (TaE) framework to address shuffled LT-CIL challenges, allowing adaptive parameter updates while significantly reducing the expandable model memory size.
    \item To further encourage class-specific feature representation, 
    we propose a Centroid Enhanced (CEd) method to guide the update of these task-aware parameters and jointly address the challenge of low discriminability of tail class features in LT-CIL.
    \item The method we proposed achieves state-of-the-art performance on all 12 benchmarks. Notably, our method achieves an average accuracy surpassing the \textit{SOTA} by 3.69$\%$ on CIFAR-100 and 2.25$\%$ on ImageNet100 under the B0-10 steps with $\rho=0.1$. 
\end{itemize}

\section{Related Work}
\subsection{Class-incremental Learning}
Recent CIL algorithms can be roughly divided into model-centric and algorithm-centric methods \cite{deepCILsurvey}. Within the model-centric methods, dynamic network algorithms have demonstrated commendable performances. Yan et al extend a new network backbone when faced with a new task, subsequently aggregating it at the feature level with a larger classifier \cite{DER}. Wang et al. contend that not all features are necessarily effective and, building on the foundation set by DER, employ knowledge distillation for model reduction \cite{Foster}. Zhou et al. decouples the intermediate layers of the network. Extending the deep network, reduces memory overhead, addressing the issue of excessive memory budget cost by dynamic network expansion \cite{memo}. On the other hand, within algorithm-centric methods, knowledge distillation methods stand out in terms of efficacy. Li et al were the first to employ knowledge distillation in CIL, establishing a regularization term via knowledge distillation to counteract forgetting \cite{lwf}. Rebuffi et al. enhance LWF by utilizing an exemplar set and introducing a novel selection method called \textit{herding} \cite{icarl}. Wang et al. \cite{CILTCSVT} proposed a novel Semantic knowledge-guided ciL framework to address the inter-class confusion problem in incremental learning. This framework utilizes semantic knowledge extracted from class labels to construct two inter-class relation graphs encompassing all encountered old and new classes.

\subsection{Long-tailed Learning}
The phenomenon of long-tail data distribution is pervasive in the real world, and the challenges of long-tail learning have been studied extensively. Most existing works can be categorized into class rebalancing, information augmentation, and model improvement. Class rebalancing focuses on techniques such as resampling and reweighting. Lin et al. investigated the difficulty of class prediction as a method for re-weighting \cite{focal}. Cao et al. impose variable margin factors for distinct classes, determined by their training label occurrences, prompting tail classes to possess broader margins \cite{LDAM}. Beyond basic image manipulations such as rotation and cropping, transfer learning has also emerged as a focal point in recent research on data augmentation. Wang et al introduced a MetaNetwork that transforms few-shot model parameters to many-shot ones \cite{MetaModelNet}.  Liu et al. focus on head-to-tail transfer at the classifier tier \cite{GIST}. Yin et al. leverage head-class variance insights to bolster feature augmentation for tail classes, aiming for greater intra-class variance in tail-class features \cite{FTL}. In the model improvement section, the most popular research focuses on training by decoupling the feature extractor and the classifier. Kang et al. first introduced the concept of a two-stage decoupled training scheme \cite{decoupling}. Chu et al. introduced an innovative classifier re-training method using tail-class feature augmentation \cite{OFA}. 

\subsection{Long-tail Class-incremental Learning}
Recently, Liu et al. \cite{LTCIL} incorporated long-tail distributions into class incremental learning (LT-CIL), drawing inspiration from the learnable weight scaling (LWS) approach introduced by the decoupling strategy \cite{decoupling}, and demonstrated that the long-tail distribution of data not only deteriorates the performance of class incremental learning algorithms but also compromises their robustness. "Based on Liu et al.'s work, Kalla et al. \cite{GVAlign} achieve classifier alignment by leveraging global variance as an informative measure and class prototypes in the second stage. While this approach enhances the robustness of tail class features, aligning all classes using the feature space of the largest class may result in an inadequate distribution of the overall feature space, thereby affecting the model’s plasticity. Therefore, the CEd proposed in this paper can adaptively adjust the feature space size for each class during the training process. Wang et al.\cite{ISC} constructed two parallel spaces sub-prototype and reminiscence spaces—to learn robust representations and mitigate forgetting in LT-CIL.

\section{Method}
\begin{figure}[t]
\centering
\includegraphics[width=0.98\textwidth]{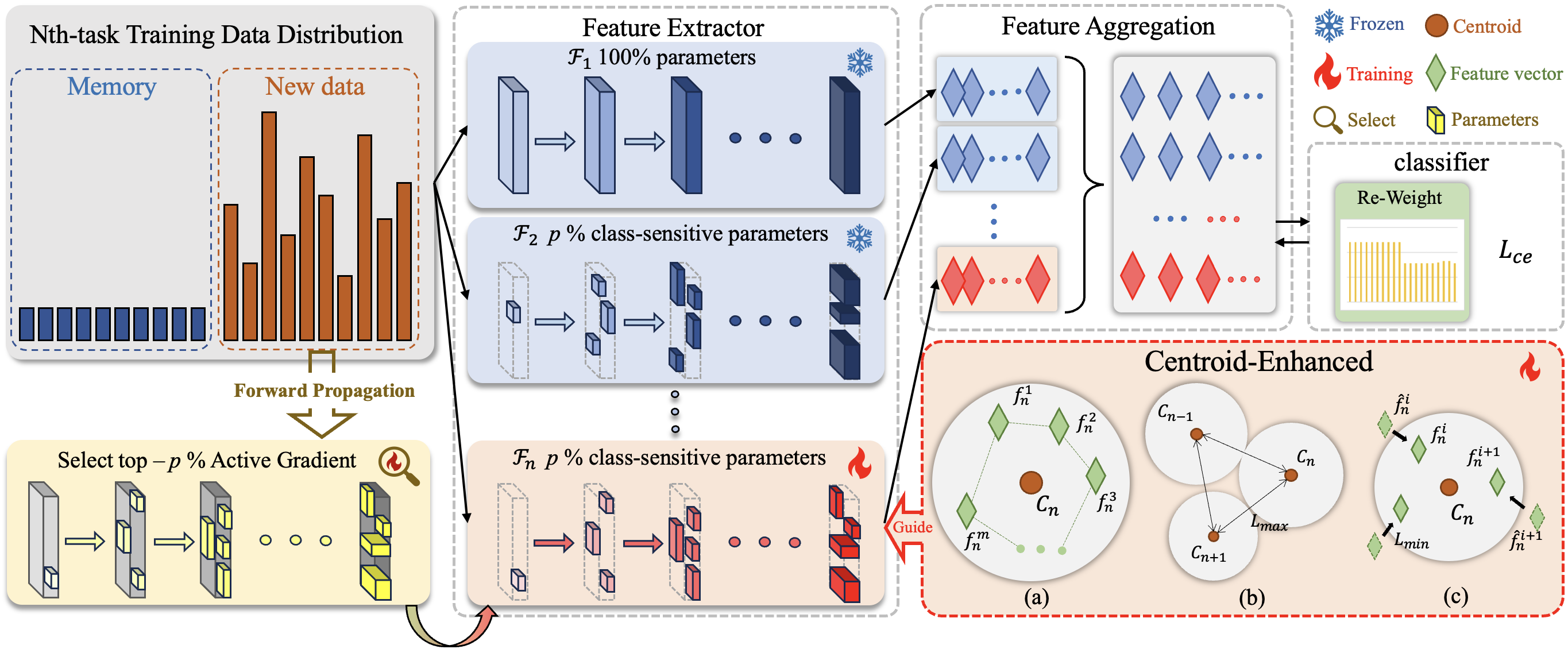}
\caption{The overview of TaE. During training for the $n$ task, the training set undergoes multiple forward passes on the model from the previous round. It selects the most sensitive $p\%$ of gradients associated with these parameters, expands these selected parameters, and freezes the others. The learning process is guided by the Centroid-Enhanced (CEd) method: (a) each class learns a centroid updated throughout training; (b) centroids between different classes remain distant; (c) features within the same class converge towards the centroid. This process is governed by the $\mathcal{L}_{max-min}$ loss. The Re-weight strategy trains classifier learning.}
\label{overview}
\end{figure}

\subsection{Preliminary}
To formalize, we denote the sequence of $T$ training tasks as $\left \{  {\mathcal{D}^{1},\mathcal{D}^{2}, \cdots, \mathcal{D}^{T}}\right \}$, where each task $\mathcal{D}^{t}$ is characterized by $\left \{ \left ( \mathbf{x}_{i}^{t},{y}_{i}^{t}    \right )  \right \}_{i=1}^{{n}_{t}}$. Here, $\mathbf{x}_{i}^{t} \in \mathbb{R}^{D} $ represents an instance belonging to class ${y}_{i}\in {Y}_{t}$, and ${Y}_{t}$ denotes the label space of task $t$. Importantly, the label spaces for different tasks are mutually exclusive. During the training of task $t$, only data from $\mathcal{D}^{t}$ is accessible. The model's performance is evaluated across all observed class labels, defined as $\mathcal{Y}_{t}={Y}_{1}\cup \cdots {Y}_{t}$ \cite{deepCILsurvey}.

The Exemplar Set is an auxiliary collection of instances from previous tasks, denoted as $\mathcal{E}=\left \{ \left ( \mathbf{x}_{j},{y}_{j}  \right )  \right \}_{j=1}^{M}$, where ${y}_{j}\in \mathcal{Y}_{t-1} $. This set, combined with the current dataset, i.e., $\mathcal{E} \cup \mathcal{D}^{t}$, facilitates model updates within each task \cite{memo}. One prominent method for choosing exemplars that is widely used is herding, as suggested in \cite{icarl}.

The cross-entropy loss $ \mathcal{L}_{ce} $ at task $t$ is define as:
\begin{equation}
\label{CEloss}
\mathcal{L}_{ce, t} (\mathbf{x},y) = -\frac{1}{\left | \mathcal{D}^{t}  \right |+\left | \mathcal{E}  \right |  } \sum_{(\mathbf{x} ,y)\in \mathcal{D}^{t}\cup \mathcal{E}}  \mathbf{y}   \cdot \log(p_{1:t}(\mathbf{x} )) 
\end{equation}
where $\mathbf{y}$ is a one-hot vector in which the position of the ground-truth label is 1, and $p_{1:t}(\mathbf{x})$ is a vector with the probability predictions for image $\mathbf{x}$ overall seen classes.

An overview of our approach is illustrated in Fig.\ref{overview}.

\subsection{Extraction and Expansion of Task-Aware Parameters}

Recent research has observed that the pre-trained backbone exhibits distinct feature patterns at different locations \cite{featurepattern,featurepattern2}. When confronted with downstream tasks, it is unnecessary to completely retrain all parameters. Instead, adjusting the domain-specific parameters based on the features of different task data proves more effective than a full retraining approach \cite{sensitivity}. Inspired by this, the domains between new and old tasks are closely related (e.g., both involving image classification tasks) in CIL, and leveraging the characteristics of dynamic network methods, we treat the model learned in the $t$-1-th task as the "pre-trained network" for the $t_{th}$ task. Consequently, we select task-aware parameters for expansion and freeze the majority of parameters to mitigate forgetting.

Formally, given the training dataset $D^{t}$ for the $t_{th}$ task, and the $t-1_{th}$ task model $\mathcal{F}_{t-1}$. We initially pass $D^{t}$ through the $\mathcal{F}_{t-1}$. Through multiple iterations involving forward propagation, we compute the gradient of each parameter and accumulate the changes in gradients over several loops, and the parameters are sorted based on their magnitudes. Finally, the top $p$$\%$ (e.g., 5$\%$, 10$\%$, 20$\%$, 30$\%$) of the most sensitive parameters are selected. The algorithmic process is illustrated in Algorithm.\ref{Select}.

\begin{algorithm}[H]
\small
\caption{Select Top-$p\%$ Most Sensitive Parameters}
\label{Select}
\begin{algorithmic}[1]
\State \textbf{Input:}  ${t}^{th}$ task data $\mathcal{D}^t$, Top percentage $p$, The 
number of forward iteration $Z$
\State \textbf{Require:} $\theta$ // parameter of $\mathcal{F} _{t-1}$, 
\State \textbf{Output:} Top $p\%$ sensitive parameters
\For{$i = 1$ to $n$}
    \For{$(\mathbf{x}_i^t, y_i^t) \in \mathcal{D}^t$}
        \State Perform forward pass: $\mathcal{F} _{t-1}(\mathbf{x}_i^t) \rightarrow \hat{y}_i$
        \State Compute $\mathcal{L}_{ce} $ by Eq \ref{CEloss}
        \State Compute gradients: $\nabla \theta_i = \frac{\partial \mathcal{L}_{ce}}{\partial \theta}$
        \State Accumulate gradients: $\text{G}_\theta = \text{G}_\theta + \nabla \theta_i$
    \EndFor
\EndFor
\State Average accumulated gradients: $\bar{  \text{G}_\theta} = \frac{\text{G}_\theta}{Z}$

\State Sort parameters based on gradient magnitude
\State $\text{$\theta_{select}$} = \text{Top}(\text{G}_\theta, p)$
\State \textbf{return} $\theta_{select}$

\end{algorithmic}
\end{algorithm}

\subsection{Centroid-Enhanced Method}
Due to the issue of long-tailed distribution, the model often exhibits a bias towards the head classes, leading to a lack of learning from tail class samples. This results in the features of the tail class being less discriminative. This problem with tail classes becomes more pronounced in CIL. To tackle this issue, we introduce a Feature Isolation Strategy designed to steer the model towards learning class-specific features. In detail, we set a learnable centroid for each class of the new task and introduced a centroid bank to store centroids of all seen classes. During training, we freeze centroids of old classes in the centroid bank, learn centroids of new classes, and employ $\mathcal{L}_{min-max}$ to isolate feature spaces among different classes. We dynamically adjust the size of class feature spaces, where $\mathcal{L}_{min}$ encourages intra-class feature cohesion, and $\mathcal{L}_{max}$ promotes inter-class feature separation \cite{gas,FSCIL,gasCentroid}.

For computing the similarity between feature vectors, we opt to use cosine similarity to measure distance \cite{cosine}, which is thus: $\cos(u,v)=\frac{u \cdot v}{\left \| u \right \| \left \| v \right \| } $ where $u$ and $v$ are two vectors, $\cdot $ represents the dot product, and $\left \| \cdot  \right \|  $ is the $ L_{2}$ norm.

The \textit{min} loss ensures that features from samples within the same class gravitate towards their corresponding class centroid. It can be mathematically represented as:
\begin{equation}
\label{minloss}
\mathcal{L}_{min,t} = -\frac{1}{|\mathcal{D}^{t}|}\sum_{i=1}^{|\mathcal{D}^{t}|} \cos (f(\mathbf{x}_{i}^{t}),c_{y_{i}^{t}})
\end{equation}
where $f(\mathbf{x}_{i}^{t})$ is the features of sample $\mathbf{x}_{i}^{t}$ fitted by the model, $c_{y_{i}^{t}}$ is the centroid of label $y_{i}^{t}$.

The max loss ensures that centroids of different classes are distinct and separated in the feature space. The loss is defined as:
\begin{equation}
\label{maxloss}
\mathcal{L}_{max,t} = \frac{1}{\mathcal{K}_{t}(\mathcal{K}_{t}-1) }\sum_{j=1}^{\mathcal{K}_{t}} \sum_{k\ne j}^{\mathcal{K}_{t}}\cos (c_{j}^{t},c_{k}^{t}) 
\end{equation}
where $\mathcal{K}_{t}$ is the number of classes in task $t$.

The overall loss function at task $t$ is defined as:
\begin{equation}
\label{allloss}
\mathcal{L}_{all,t} = \gamma_{1}\mathcal{L}_{ce,t} + \gamma_{2}\mathcal{L}_{min,t} +\gamma_{3}\mathcal{L}_{max,t}
\end{equation}
where $\gamma_{1}$, $\gamma_{2}$, and $\gamma_{3}$ are the hyper-parameter to control the effort of these losses. The details of setting different hyperparameters and comparing experimental results are provided in Sec.\ref{sec: Hyper-parameter}.

\subsection{Reweight strategy}
For the training set of a single task, $\mathcal{E} \cup \mathcal{D}^{t}$, with a total of $\mathcal{K}_t$ classes and $i$-th class having a quantity of $k^{t}_i$ samples. The effective weight for each class is defined as:
\begin{equation}
\label{effective weight}
w_{i} = \frac{1-\beta }{1-\beta^{k^{t}_{i}}}  
\end{equation}
where $\beta$ is a hyperparameter that controls the weights of different classes, and $w_i$ represents the weight for the $i$-th class. Based on our experiments, we found that setting $\beta$ to 0.95 yields the best results.

\section{EXPERIMENT}
\label{experiment}
\subsection{Experiment Setup and Implementation Details}
\textbf{Datasets.} CIFAR-100 \cite{cifar} is composed of 32x32 pixel color images classified into 100 classes. The dataset comprises 50,000 training images, with 500 images per class, and 10,000 evaluation images, featuring 100 images per class. ImageNet-100 \cite{ImageNet} is curated by selecting 100 classes from the ImageNet-1000 dataset, which includes 100,000 training images, with 1,000 images per class, and 5,000 evaluation images with 50 images per class.

\textbf{Datasets Protocol.}: We follow the protocol introduced in \cite{icarl}, which involves training all 100 classes through multiple splits, encompassing 5, 10, and 20 incremental steps while maintaining a fixed memory size of 2,000 exemplars across batches. \textit{B0} indicates that the model is trained from sketch rather than pre-trained on an amount of data before incremental learning.

\begin{table}[t]
\caption{Imbalanced Ratio Protocol, H-T-M means the number of head, tail, and memory, k means 1000.}\label{Imbalanced Ratio}
\centering
\begin{tabular}{@{\extracolsep\fill}lccc}
\toprule%
& \multicolumn{1}{c}{$\rho = 0.1 $} & \multicolumn{1}{c}{$\rho = 0.05 $} & \multicolumn{1}{c}{$\rho = 0.01 $} \\\cmidrule{2-4}%
Dataset & H-T-M & H-T-M & H-T-M \\
\midrule
CIFAR100-LT & 500-50-20&	500-25-20 &	500-5-5\\
ImageNet100-LT	& 1k-100-20	& 1k-50-20	&1k-10-10\\
\bottomrule
\end{tabular}
\end{table}

\textbf{Imbalanced Ratio Protocol.}  We adopt three protocols for the head-to-tail sample count ratio in the long-tailed distribution: $\rho$ = 0.1, $\rho$ = 0.05, and $\rho$ = 0.01. In Tab.\ref{Imbalanced Ratio}, the details of three protocols combined with two datasets are presented, including the sample counts for head and tail classes, as well as the sample quantities for each category stored in the memory.

\textbf{Metrics.} In CIL methods, we use standard metrics to evaluate performance: "Last" refers to the accuracy on the most recent task, and "Avg" refers to the Average Accuracy, which calculates the accuracy across all observed classes\cite{CILsurveyTPAMI}. Let $a_{i,j}$ be the accuracy of the model on the test set of task $j$ after training from task 1 to task $i$. The Average Accuracy can be calculated as follows:
\begin{equation}
\label{metrics}
Avg(A_{T}) = \frac{1}{T} \sum_{j=1}^{T} a_{T,j}   
\end{equation}

\textbf{Comparison Methods.} We compare our method with existing CIL methods: EWC\cite{ewc}, LwF\cite{lwf}, iCaRL\cite{icarl}, Foster\cite{Foster}, MEMO\cite{memo}, DER\cite{DER}, and LT-CIL methods: UCIR\cite{ucir} and PodNet\cite{podnet} combined with LWS\cite{LTCIL} and GVAlign\cite{GVAlign}. It is important to note that,  where MEMO only extends the backbone by 30$\%$, we set the proportion of TaE extension to $p=30\%$ when compared with other methods.

\textbf{Implementation Details.} We implement our method using PyTorch \cite{pytorch} and conduct our experiments on two NVIDIA GeForce RTX 3090 GPUs. We applied a long-tailed transformation to both CIFAR-100 and ImageNet100-LT. The experiments leveraged the public implementations of existing CIL methods within the PyCIL framework \cite{pycil}. For CIFAR-100, we employed ResNet32 \cite{resnet}, with an initial learning rate set at 0.1, which was reduced by a factor of 10 at epochs 80, 120, and 150 (out of a total of 170 epochs). For ImageNet100-LT, we utilized ResNet18 with the same learning rate adjustments. In these experiments, we choose exemplars for memory based on the herding selection strategy \cite{herding}, in line with previous studies \cite{icarl}.

\subsection{Evaluation on CIFAR100-LT}
\begin{figure}[t]
\centering
\includegraphics[width=1.0\textwidth]{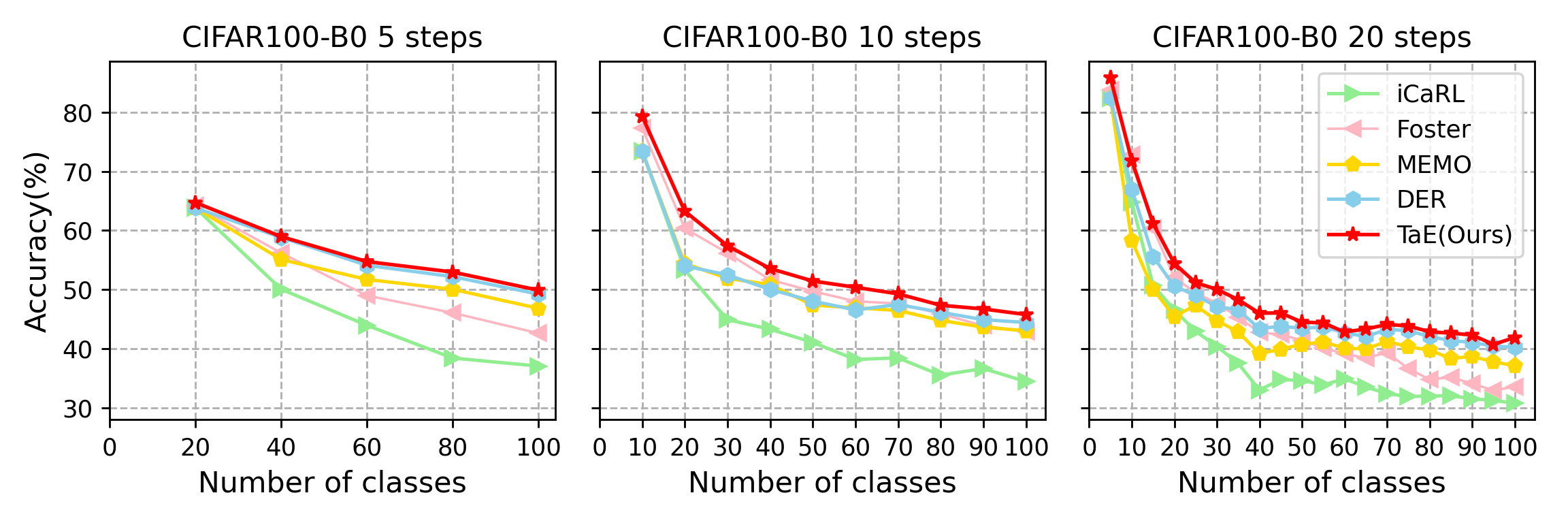}
\caption{The performance for each step. The model is trained on CIFAR100-LT $\rho$ = 0.1 with three Datasets Protocols.}
\label{comparison_cifar}
\end{figure}
\begin{table}[t]
\caption{Performance comparison of typical CIL algorithms on CIFAR100-LT B0-10steps benchmarks, evaluated across three Imbalanced Ratio Protocols.}\label{cifar100-ratio}
\centering
\begin{tabular*}{\textwidth}{@{\extracolsep\fill}lcccccc}
\toprule%
& \multicolumn{2}{c}{$\rho = 0.1 $} & \multicolumn{2}{c}{$\rho = 0.05 $} & \multicolumn{2}{c}{$\rho = 0.01 $} \\\cmidrule{2-3}\cmidrule{4-5}\cmidrule{6-7}%
Method & Last & Avg & Last & Avg & Last & Avg \\
\midrule
Finetune(baseline) & 7.8 & 22.42	&7.57	&20.98	&5.83	&15.36\\
EWC\cite{ewc} & 8.88	& 22.37	&8.45	&21.65	&5.63	&14.92\\
LwF\cite{lwf} & 16.82	& 33.51	&16.94	&28.45	&11.56	&20.39\\
iCaRL\cite{icarl} & 34.43	& 44.82	&33.63	&41.64	&14.32	&22.68\\
Foster\cite{Foster} &42.89	&53.82	&39.86	&47.28	&15.17 	&21.72\\
MEMO\cite{memo}	& 43.02	&50.94	&40.08	&46.31	&21.0	&26.39\\
DER\cite{DER} &44.43	&50.74	&39.24	&44.37	&22.89	&28.44\\
\midrule
TaE($p$=30$\%$)	& \textbf{46.84}	&\textbf{55.04}	&\textbf{40.87}	&\textbf{48.29}	&\textbf{23.54}	&\textbf{29.69}\\
\bottomrule
\end{tabular*}
\end{table}
\textbf{Quantitative Results} Tab.\ref{cifar100-ratio} and Tab.\ref{comparison_cifar} summarize the results of the CIFAR-100 benchmark tests. The experimental results demonstrate that, under similar parameter expansion, TaE's task-aware parameter expansion is more suitable for LT-CIL scenarios. The experimental results indicate that our approach outperforms other methods across all six settings in the CIFAR100-LT dataset, which substantiates the strong robustness of our method.

\subsection{Evaluation in ImageNet100-LT}
\begin{figure}[t]
\centering
\includegraphics[width=1.0\textwidth]{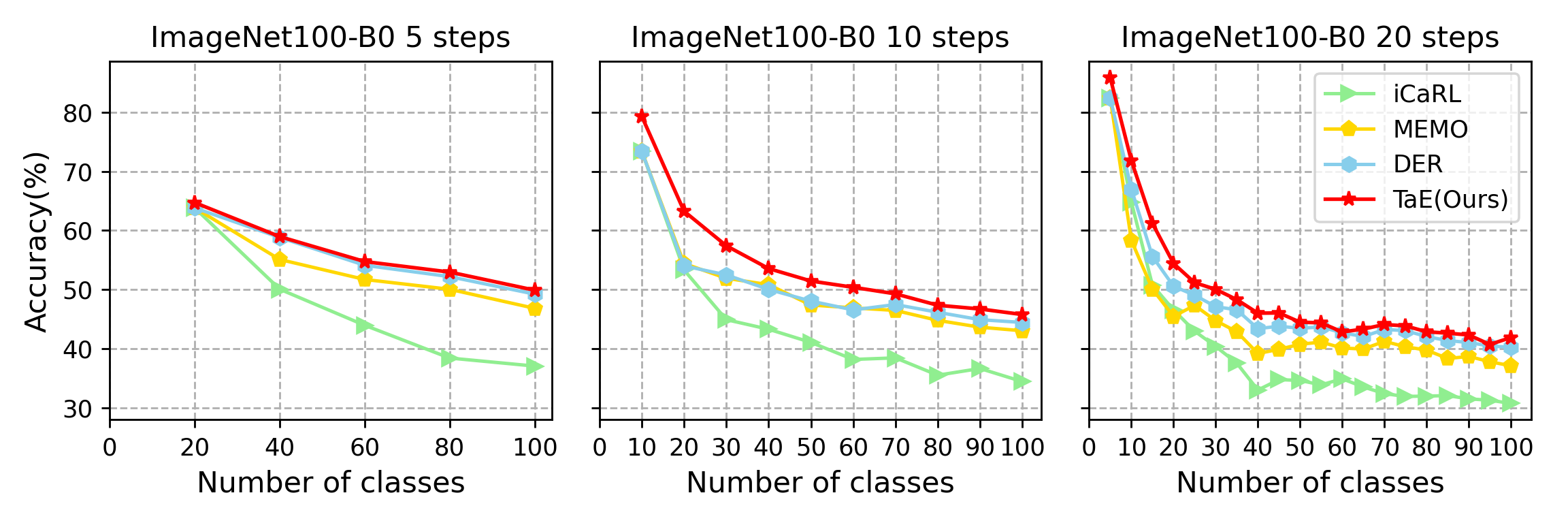}
\caption{The performance for each step. The model is trained on ImageNet100-LT B0-10steps $\rho$ = 0.1.}
\label{comparison_ImageNet}
\end{figure}
\textbf{Quantitative Results} Fig.\ref{comparison_ImageNet} illustrates the accuracy of different methods at each incremental stage, revealing our approach consistently outperforms others. Tab.\ref{ImageNet100-LT-ratio} presents the performance of different methods under three distinct long-tail ratio settings, with TaE consistently demonstrating the best performance across all scenarios.

\begin{table}[t]
\caption{Performance comparison of typical CIL algorithms on ImageNet100-LT B0-10steps benchmarks, evaluated across three Imbalanced Ratio Protocols.}\label{ImageNet100-LT-ratio}
\centering
\begin{tabular*}{\textwidth}{@{\extracolsep\fill}lcccccc}
\toprule%
& \multicolumn{2}{c}{$\rho = 0.1 $} & \multicolumn{2}{c}{$\rho = 0.05 $} & \multicolumn{2}{c}{$\rho = 0.01 $} \\\cmidrule{2-3}\cmidrule{4-5}\cmidrule{6-7}%
Method & Last & Avg & Last & Avg & Last & Avg \\
\midrule
Finetune(baseline) & 8.9 & 23.62	&9.12	&21.24	&6.10	&15.77\\
EWC\cite{ewc} & 10.83	& 24.74	&9.32	&23.02	&7.06	&17.17\\
LwF\cite{lwf} & 19.62	& 38.76	&18.04	&34.67	&14.18	&25.11\\
iCaRL\cite{icarl} & 33.72	& 44.40	&32.98	&42.51	&15.78	&23.49\\
Foster\cite{Foster} &49.08	&59.15	&44.07	&50.40	&18.10 	&23.12\\
MEMO\cite{memo}	& 48.78	&54.73	&44.10	&50.40	&25.72	&30.47\\
DER\cite{DER} &50.32	&57.48	&41.22	&50.69	&26.82	&31.22\\
\midrule
TaE($p$=30$\%$)	& \textbf{52.86}	&\textbf{59.73}	&\textbf{44.69}	&\textbf{55.28}	&\textbf{29.98}	&\textbf{33.75}\\
\bottomrule
\end{tabular*}
\end{table}

\subsection{Comparison with LT-CIL Methods}
Tab.\ref{cifar100-LWS-GVAlign} presents a performance comparison of our method and three existing LTCIL methods (LWS \cite{LTCIL} and GVAlign \cite{GVAlign}) under the same experimental settings. As the table shows, TaE consistently demonstrates superior efficacy across varied experimental conditions.

\begin{table}[t]
\caption{Performance comparison of LTCIL algorithms on CIFAR100-LT and ImageNet100-LT B0-5steps and 10steps with $\rho = 0.1$ benchmarks.}\label{cifar100-LWS-GVAlign}
\centering
\begin{tabular*}{\textwidth}{@{\extracolsep\fill}lcccccc}
\toprule%
& &\multicolumn{2}{c}{CIFAR100-LT} & \multicolumn{2}{c}{ImageNet100-LT} \\\cmidrule{3-4}\cmidrule{5-6}%
Method & Conference & B0-10steps & B0-5steps & B0-10steps & B0-5steps \\
\midrule
UCIR(+LWS)\cite{LTCIL} &ECCV2022 & 39.40 & 39.00&	49.42 	&47.96	\\
PodNet(+LWS)\cite{LTCIL} &ECCV2022	& 36.37	& 37.03	&49.75	&49.51	\\
UCIR(+GVAlign)\cite{GVAlign} &WACV2024 & 42.80 &41.64	&50.69	&47.58\\
PodNet(+GVAlign)\cite{GVAlign} &WACV2024 & 42.72	& 41.61	&52.01	&50.81\\
\midrule
TaE($p$=30$\%$) & ACCV2024 	& \textbf{54.97}	&\textbf{55.04}	&\textbf{59.73}	&\textbf{64.48}\\
\bottomrule
\end{tabular*}
\end{table}

\subsection{Ablation Study}
To validate the effectiveness of each component in TaE, we conducted an ablation study on CIFAR100-LT B10-10steps with $\rho$ = 0.1. As shown in Tab.\ref{ablation}, both the average and final accuracy progressively improved as we added more components. We observe that, building upon TaE($30\%$), the inclusion of Re-weight results in an average accuracy improvement of $1.16\%$, and further incorporating $\mathcal{L}_{max-min}$ leads to an additional average accuracy enhancement of $1.35\%$.
\begin{table}[t]
\caption{Ablations of key components in TaE($30\%$). We report the average and last accuracy on CIFAR100-LT B0-10 steps. RW means Re-Weight strategy.}\label{ablation}
\centering
\begin{tabular}{@{\extracolsep\fill}lccccc}
\toprule%
RW & $\mathcal{L}_{max-min}$ & Last & Avg  \\
\midrule
\XSolidBrush & \XSolidBrush  & 45.03 & 51.92 \\
\checkmark & \XSolidBrush  & 45.79 & 53.08 \\
\checkmark & \checkmark  & 46.84 & 55.04 \\
\bottomrule
\end{tabular}
\end{table}

\begin{figure}[t]
  \centering
  \begin{subfigure}[b]{0.48\textwidth}
    \includegraphics[width=\textwidth]{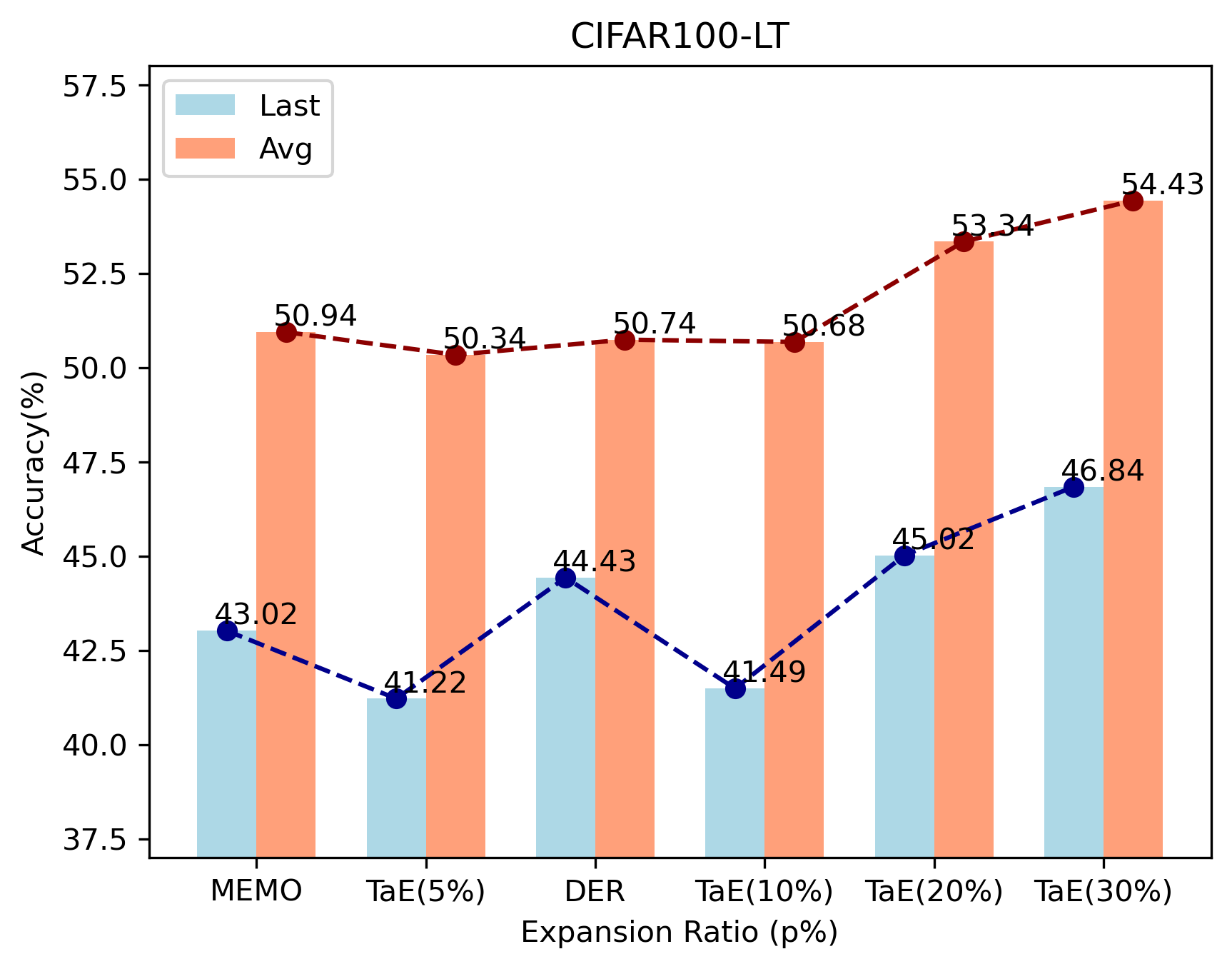}
    \caption{}
    \label{histogram_cifar}
  \end{subfigure}
  \begin{subfigure}[b]{0.48\textwidth}
    \includegraphics[width=\textwidth]{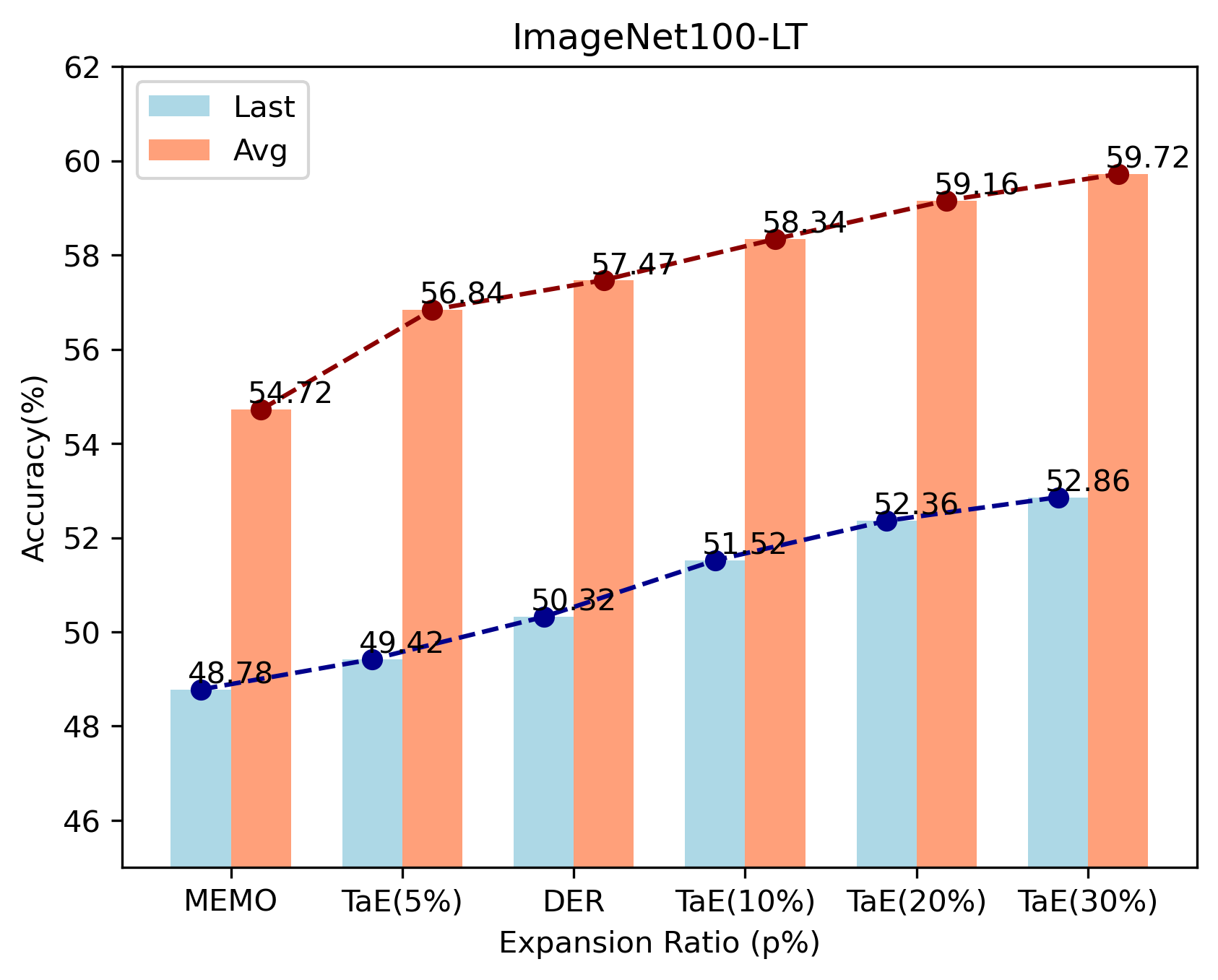}
    \caption{}
    \label{histogram_ImageNet}
  \end{subfigure}
  \caption{The Last and Average Accuracy. We compare our model($p$ = 5$\%$, 10$\%$, 20$\%$, 30$\%$) with \textit{SOTA} CIL methods (DER \cite{DER} and MEMO \cite{memo}), and the model trained on the experiment CIFAR100-LT and ImageNet100-LT B0-10steps with $\rho = 0.1$.}
\end{figure}

\subsection{Discussion on the Selection of Parameter Ratios}
\label{sec: parameter ratio}

We conducted experiments on the ImageNet100-LT and CIFAR100-LT datasets using the B0-10steps benchmark with $p=0.1$, comparing the performance of TaE ($5\%, 10\%, 20\%, 30\%$) with MEMO and DER. When training the model on ImageNet100-LT, as illustrated in Fig.\ref{histogram_ImageNet}, the performance of TaE increases with higher values of $p$. TaE surpasses MEMO at $p=5\%$ and exceeds DER at $p=10\%$. TaE($30\%$) outperforms MEMO in terms of both final accuracy and average accuracy, achieving superiority of $4.08\%$ and $5.00\%$, respectively. Moreover, compared to DER, TaE($30\%$) exhibits a superiority of $2.54\%$ in final accuracy and $2.25\%$ in average accuracy.

However, as illustrated in Fig.\ref{histogram_cifar}, when training the model on CIFAR100, TaE only surpasses DER and MEMO at $p=20\%$. TaE($30\%$) surpasses MEMO in both final accuracy and average accuracy, demonstrating advantages of $3.82\%$ and $3.39\%$, respectively. Additionally, in comparison to DER, TaE($30\%$) demonstrates a superiority of $2.41\%$ in final accuracy and $3.71\%$ in average accuracy.

\subsection{Discussion on Hyper-parameter}
\label{sec: Hyper-parameter}
\begin{figure}[t]
\centering
\includegraphics[width=0.6\columnwidth]{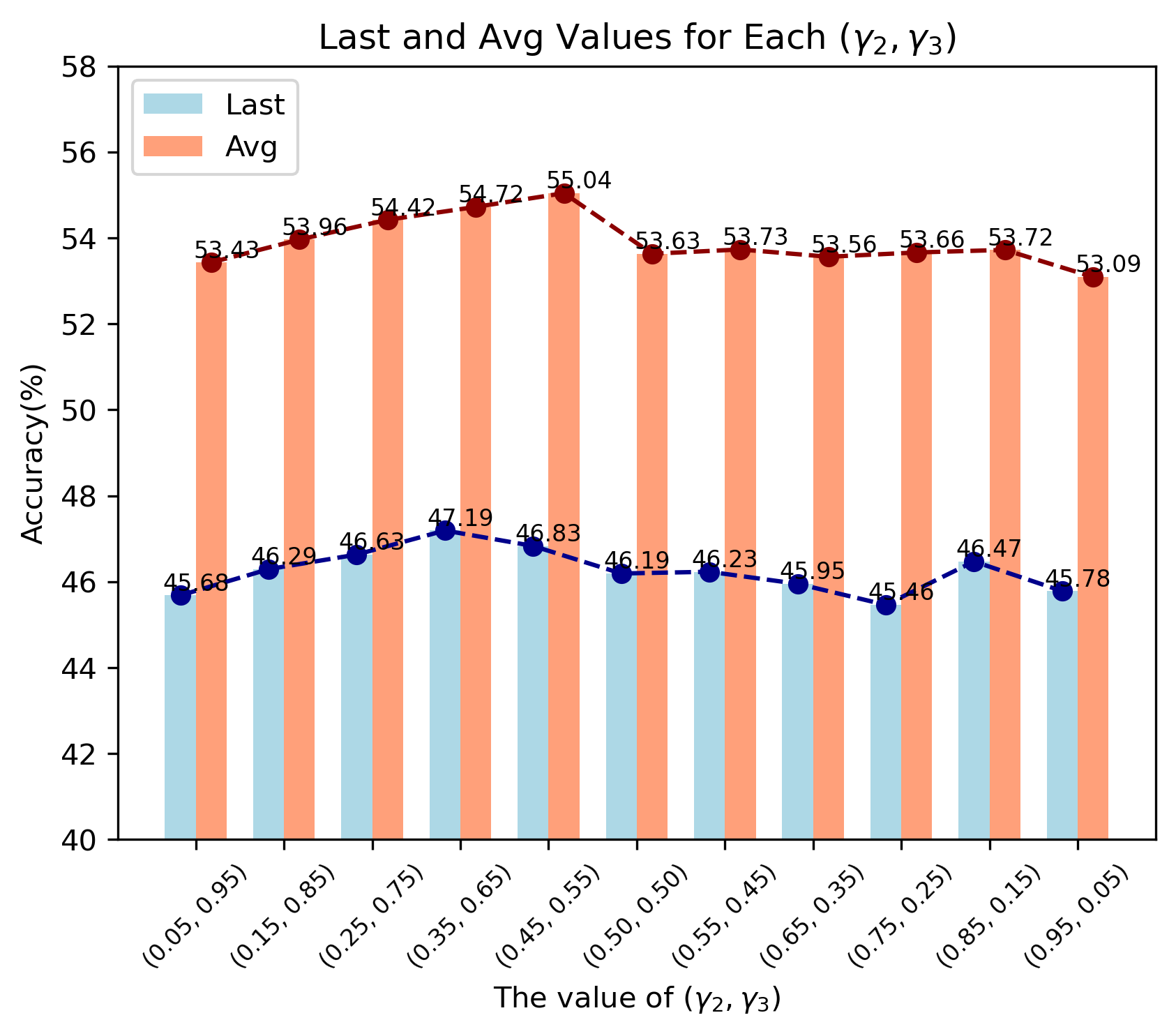}
\caption{The Last and Average Accuracy. We compare our model($p$ = 30$\%$) across various combinations of $\mathcal{L}_{max}$ and $\mathcal{L}_{min}$ and the model trained on the experiment CIFAR100-LT B0-10steps with $\rho = 0.1$.}
\label{histogram_gamma}
\end{figure}
We conducted a sensitivity analysis on the hyperparameters in the Eq.\ref{allloss}. For simplicity, we set $\gamma_{1}$ to 1 and explored various combinations of  $\gamma_{2}$ and  $\gamma_{3}$ to assess the importance of $\mathcal{L}_{min}$ and $\mathcal{L}_{max}$. From the experimental results, it is evident that the performance of all hyperparameter combinations surpasses that of the State-of-the-Art method, confirming the robustness of TaE. The best performance is achieved when the values of $\gamma_{2}$ and $\gamma_{3}$ are close, and the model slightly leans towards emphasizing inter-class distances. The results show a final accuracy of 47.19$\%$ and an average accuracy of 54.72$\%$, surpassing the SOTA by 2.76$\%$ and 3.98$\%$, respectively.

\subsection{Effectiveness of Centroid-Enhanced Method}
\begin{table}[!t]
\caption{Performance comparison of four CIL algorithms combined with CEd on the CIFAR100 $\rho$ = 0.1 benchmarks, evaluated across three Datasets Protocols.}
\centering
\begin{tabular*}{\textwidth}{@{\extracolsep\fill}lcccccc}
    \toprule
    & \multicolumn{2}{c}{B0-5 steps} & \multicolumn{2}{c}{B0-10 steps} & \multicolumn{2}{c}{B0-20 steps} \\\cmidrule{2-3}\cmidrule{4-5}\cmidrule{6-7}%
    Method & Last & Avg & Last & Avg & Last & Avg \\
    \midrule
    LwF\cite{lwf} & 8.15	& 23.12	& 17.56	& 32.87	& 27.02	& 40.09	\\
    w CEd	& \textbf{8.86}	& \textbf{24.23}	& \textbf{17.84}	& \textbf{33.79}	& \textbf{30.99}	& \textbf{41.22}\\
    \midrule
    iCarL\cite{icarl} & 37.05&	46.51	&34.13	&43.47	&30.74 &39.59	\\
    w CEd	& \textbf{43.76}	&\textbf{52.35}	&\textbf{41.24}	&\textbf{51.95}	&\textbf{33.35}	&\textbf{43.51}\\
    \midrule
    DER\cite{DER} &46.26	&52.91	&44.43	&50.74	&39.70	&46.31\\
    w CEd	& \textbf{48.51}	&\textbf{56.50}	&\textbf{45.07}	&\textbf{53.23}	&\textbf{40.05} 	&\textbf{47.52}\\
    \midrule
    Foster\cite{Foster} & 41.82	&51.46	&40.44	&51.23	&35.03	&45.51\\
    w CEd	& \textbf{42.08	}&\textbf{51.73}	&\textbf{40.98}	&\textbf{52.67}	&\textbf{36.97}	&\textbf{48.12}\\
    \bottomrule
    \end{tabular*}
\label{cifar100_ce}
\end{table}

The centroid-enhanced(\textbf{CEd}) method can be applied to most "training from sketch" CIL methods. In this section, we will validate the effectiveness of the centroid set in various experimental settings. In Tab.\ref{cifar100_ce}, we present the experimental results on the CIFAR100 with $\rho$ = 0.1. We applied CEd to four representative CIL methods. Our experiments were conducted under three protocols: B0-5steps, B0-10steps, and B0-20steps.  

The experimental results reveal that the inclusion of the centroid set enhances the performance of four representative CIL methods. We can observe improvements across various benchmarks for all four CIL methods after the incorporation of the CEd approach. Particularly, in the experimental setup of iCaRL with B0-10 steps, the final accuracy and average accuracy in the last stage increased by $7.11\%$ and $8.48\%$, respectively. The average accuracy even surpasses that of DER and Foster under the same experimental conditions. In other benchmarks, the improvement of iCaRL is notably pronounced. The above experiments demonstrate that our proposed centroid method effectively strengthens the ability of CIL to handle LT-CIL.

\section{Conclusions}
In our study, we have developed a task-aware expandable framework to tackle the challenges associated with long-tail class incremental learning. We identify the most gradient-sensitive parameters in the final model after each training task by utilizing forward propagation to achieve this. We then expand and train these parameters to improve the model's representational capabilities without causing memory overflow. Additionally, we have introduced a centroid-enhanced method to facilitate the training of task-aware parameters. This method can be adapted to various class-incremental learning methods. Our approach was rigorously evaluated through comprehensive experimentation. The experimental results demonstrate the superiority of our method. We believe that our work will lay the groundwork for future research in the field of long-tailed class-incremental learning.

\begin{credits}
\subsubsection{\ackname} The paper is supported by the National Natural Foundation Science of China (62101061).
\subsubsection{\discintname} The authors have no competing interests to declare that are relevant to the content of this article.
\end{credits}

\bibliographystyle{splncs04}
\bibliography{main}

\end{document}